\documentstyle[aclap]{article}

\newtheorem{definition}{Definition}[section]

\newcommand{\topu}{\top^u}
\newcommand{\topi}{\top^i}
\newcommand{\topd}{\top^d}
\newcommand{\botu}{\bot^u}
\newcommand{\boti}{\bot^i}
\newcommand{\botd}{\bot^d}
\newcommand{\uentails}{\models^u}
\newcommand{\ientails}{\models^i}
\newcommand{\dentails}{\models^d}

\newcommand{\forallu}{\forall^{U\!t}}

\newcommand{\name}[1]{\label{#1}}
\newcommand{\sent}[1]{(\ref{#1})}

\newcounter{stce}[part]
\newcounter{next}[part]
\newcounter{nextone}[part]
\newcounter{lastone}[part]
\def\stce
    {  \refstepcounter{stce}
       \stepcounter{next}
       \stepcounter{nextone}
       \stepcounter{lastone}
       \begin{itemize}
       \item[{\bf (\thestce )}~]%
    }

\def\endstce
    {\end{itemize}}

\def\example#1{\stce#1\endstce}

\def\ul#1{$\underline{\smash{\hbox{#1}}}$}

\title{\vspace{-0.5in}A Uniform Treatment of Pragmatic Inferences in
Simple and Complex Utterances and Sequences of Utterances}
\author{Daniel Marcu and Graeme Hirst\\
Department of Computer Science \\
University of Toronto \\
Toronto, Ontario \\
Canada \hspace{5mm} M5S 1A4 \\
{\tt \{marcu,gh\}@cs.toronto.edu} \\}

\begin{document}

\maketitle
\vspace{-0.5in}
\begin{abstract}

  Drawing appropriate defeasible inferences has been proven to be one
  of the most pervasive puzzles of natural language processing and a
  recurrent problem in pragmatics.  This paper provides a theoretical
  framework, called {\em stratified logic}, that can accommodate
  defeasible pragmatic inferences. The framework yields an algorithm
  that computes the conversational, conventional, scalar, clausal, and
  normal state implicatures; and the presuppositions that are
  associated with utterances. The algorithm applies equally to simple
  and complex utterances and sequences of utterances.

\end{abstract}

\bibliographystyle{acl}

\section{Pragmatics and Defeasibility}

It is widely acknowledged that a full account of natural language
utterances cannot be given in terms of only syntactic or semantic
phenomena. For example, Hirschberg~\shortcite{hirschberg85} has shown
that in order to understand a scalar implicature, one must analyze the
conversants' beliefs and intentions. To recognize normal state
implicatures one must consider mutual beliefs and
plans~\cite{green90}. To understand conversational implicatures
associated with indirect replies one must consider discourse
expectations, discourse plans, and discourse
relations~\cite{green92,green94}. Some presuppositions are inferrable
when certain lexical constructs (factives, aspectuals, etc) or
syntactic constructs (cleft and pseudo-cleft sentences) are used.
Despite all the complexities that individualize the recognition stage
for each of these inferences, {\em all} of them can be defeated by
context, by knowledge, beliefs, or plans of the agents that constitute
part of the context, or by other pragmatic rules.

{\em Defeasibility} is a notion that is tricky to deal with, and
scholars in logics and pragmatics have learned to circumvent it or
live with it. The first observers of the phenomenon preferred to keep
defeasibility outside the mathematical world. For
Frege~\shortcite{frege92}, Russell~\shortcite{russell05}, and
Quine~\shortcite{quine49} ``everything exists''; therefore, in their
logical systems, it is impossible to formalize the cancellation of the
presupposition that definite referents
exist~\cite{hirst91,marcu-tilburg94}. We can taxonomize previous
approaches to defeasible pragmatic inferences into three categories
(we omit here work on defeasibility related to linguistic phenomena
such as discourse, anaphora, or speech acts).

  1.  Most linguistic approaches account for the defeasibility of
  pragmatic inferences by analyzing them in a context that consists of
  all or some of the previous utterances, including the current one.
  Context~\cite{karttunen74,kay92}, procedural
  rules~\cite{gazdar79,karttunen79}, lexical and syntactic
  structure~\cite{weischedel79}, intentions~\cite{hirschberg85}, or
  anaphoric constraints~\cite{sandt92,zeevat92} decide what
  presuppositions or implicatures are projected as pragmatic
  inferences for the utterance that is analyzed. The problem with
  these approaches is that they assign a dual life to pragmatic
  inferences: in the initial stage, as members of a simple or complex
  utterance, they are defeasible.  However, after that utterance is
  analyzed, there is no possibility left of cancelling that inference.
  But it is natural to have implicatures and presuppositions that are
  inferred and cancelled as a sequence of utterances proceeds:
  research in conversation repairs~\cite{hirst94} abounds in such
  examples. We address this issue in more detail in
  section~\ref{section-seq}.

  2. One way of accounting for cancellations that occur later in the
  analyzed text is simply to extend the boundaries within which
  pragmatic inferences are evaluated, i.e., to look ahead a few
  utterances. Green~\shortcite{green92} assumes that implicatures are
  connected to discourse entities and not to utterances, but her
  approach still does not allow cancellations across discourse units.

  3.  Another way of allowing pragmatic inferences to be cancelled is
  to assign them the status of defeasible information.
  Mercer~\shortcite{mercerphd} formalizes presuppositions in a logical
  framework that handles defaults~\cite{reiter80}, but this approach
  is not tractable and it treats natural disjunction as an
  exclusive-or and implication as logical equivalence.

Computational approaches fail to account for the cancellation of
pragmatic inferences: once presuppositions~\cite{weischedel79} or
implicatures~\cite{hirschberg85,green92} are generated, they can never
be cancelled. We are not aware of any formalism or computational
approach that offers a unified explanation for the cancellability of
pragmatic inferences in general, and of no approach that handles
cancellations that occur in sequences of utterances.

It is our aim to provide such an approach here.  In doing this, we
assume the existence, for each type of pragmatic inference, of a set of
necessary conditions that must be true in order for that inference to
be triggered. Once such a set of conditions is met, the corresponding
inference is drawn, but it is assigned a defeasible status. It is the
role of context and knowledge of the conversants to ``decide'' whether
that inference will survive or not as a pragmatic inference of the
structure.  We put no boundaries upon the time when such a
cancellation can occur, and we offer a unified explanation for
pragmatic inferences that are inferable when simple utterances,
complex utterances, or sequences of utterances are considered.

We propose a new formalism, called ``stratified logic'', that correctly
handles the pragmatic inferences, and we start by giving a very brief
introduction to the main ideas that underlie it.  We give the main
steps of the algorithm that is defined on the backbone of stratified
logic. We then show how different classes of pragmatic inferences can
be captured using this formalism, and how our algorithm computes the
expected results for a representative class of pragmatic inferences.
The results we report here are obtained using an implementation
written in Common Lisp that uses Screamer~\cite{siskind93a}, a macro
package that provides nondeterministic constructs.

\section{Stratified logic}

\subsection{Theoretical foundations}

We can offer here only a brief overview of stratified logic. The
reader is referred to Marcu~\shortcite{marcu-master94} for a
comprehensive study. Stratified logic supports one type of
indefeasible information and two types of defeasible information,
namely, infelicitously defeasible and felicitously defeasible.  The
notion of infelicitously defeasible information is meant to capture
inferences that are anomalous to cancel, as in:
\example{* John regrets that Mary came to the party but she did not
  come. \name{regret-yes-cancel}}
The notion of felicitously defeasible information is meant to capture
the inferences that can be cancelled without any abnormality, as in:
\example{John does not regret that Mary came to the party because she
  did not come. \name{regret-not-cancel}}

\begin{figure}
\setlength{\unitlength}{0.012500in}%
\begingroup\makeatletter\ifx\SetFigFont\undefined
\def\x#1#2#3#4#5#6#7\relax{\def\x{#1#2#3#4#5#6}}%
\expandafter\x\fmtname xxxxxx\relax \def\y{splain}%
\ifx\x\y   
\gdef\SetFigFont#1#2#3{%
  \ifnum #1<17\tiny\else \ifnum #1<20\small\else
  \ifnum #1<24\normalsize\else \ifnum #1<29\large\else
  \ifnum #1<34\Large\else \ifnum #1<41\LARGE\else
     \huge\fi\fi\fi\fi\fi\fi
  \csname #3\endcsname}%
\else
\gdef\SetFigFont#1#2#3{\begingroup
  \count@#1\relax \ifnum 25<\count@\count@25\fi
  \def\x{\endgroup\@setsize\SetFigFont{#2pt}}%
  \expandafter\x
    \csname \romannumeral\the\count@ pt\expandafter\endcsname
    \csname @\romannumeral\the\count@ pt\endcsname
  \csname #3\endcsname}%
\fi
\fi\endgroup
\begin{picture}(105,97)(20,715)
\thinlines
\put( 30,770){\vector( 0, 1){ 25}}
\put( 90,770){\vector( 0, 1){ 25}}
\put( 30,770){\vector( 2, 1){ 54}}
\put( 90,770){\vector(-2, 1){ 54}}
\put( 30,730){\vector( 0, 1){ 25}}
\put( 90,730){\vector( 0, 1){ 25}}
\put( 30,730){\vector( 2, 1){ 54}}
\put( 90,730){\vector(-2, 1){ 54}}
\put( 85,800){\makebox(0,0)[lb]{\smash{\SetFigFont{12}{14.4}{rm}$\botd$}}}
\put( 25,800){\makebox(0,0)[lb]{\smash{\SetFigFont{12}{14.4}{rm}$\topd$}}}
\put( 85,760){\makebox(0,0)[lb]{\smash{\SetFigFont{12}{14.4}{rm}$\boti$}}}
\put( 25,760){\makebox(0,0)[lb]{\smash{\SetFigFont{12}{14.4}{rm}$\topi$}}}
\put( 25,715){\makebox(0,0)[lb]{\smash{\SetFigFont{12}{14.4}{rm}$\topu$}}}
\put( 85,715){\makebox(0,0)[lb]{\smash{\SetFigFont{12}{14.4}{rm}$\botu$}}}
\put(120,795){\makebox(0,0)[lb]{\smash{\SetFigFont{12}{14.4}{rm}{\small
Felicitously {\bf D}efeasible Layer}}}}
\put(120,755){\makebox(0,0)[lb]{\smash{\SetFigFont{12}{14.4}{rm}{\small
{\bf I}nfelicitously Defeasible Layer}}}}
\put(120,715){\makebox(0,0)[lb]{\smash{\SetFigFont{12}{14.4}{rm}{\small
{\bf U}ndefeasible Layer}}}}
\end{picture}

\caption{The lattice that underlies stratified logic}
\label{my-lattice}
\end{figure}
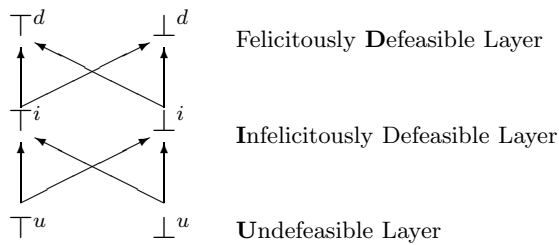

The lattice in figure~\ref{my-lattice} underlies the semantics of
stratified logic.  The lattice depicts the three levels of strength
that seem to account for the inferences that pertain to natural
language semantics and pragmatics: indefeasible information belongs to
the $u$ layer, infelicitously defeasible information belongs to the
$i$ layer, and felicitously defeasible information belongs to the $d$
layer. Each layer is partitioned according to its polarity in truth,
$\topu,\topi,\topd$, and falsity, $\botu,\boti,\botd$. The lattice
shows a partial order that is defined over the different levels of
truth. For example, something that is indefeasibly false, $\botu$, is
stronger (in a sense to be defined below) than something that is
infelicitously defeasibly true, $\topi$, or felicitously defeasibly
false, $\botd$. Formally, we say that the $u$ level is stronger than
the $i$ level, which is stronger than the $d$ level: $u\!<\!i\!<\!d$.
At the syntactic level, we allow atomic formulas to be labelled
according to the same underlying lattice. Compound formulas are
obtained in the usual way. This will give us formulas
such as $regrets^u(John, come(Mary,party)) \rightarrow
come^i(Mary,party))$, or $(\forall x)(\neg bachelor^u(x) \rightarrow
(male^d(x) \wedge adult^d(x)))$. The satisfaction relation is split
according to the three levels of truth into u-satisfaction,
i-satisfaction, and d-satisfaction:

\begin{definition} \label{xentailsfol}
Assume $\sigma$ is an  ${\cal SL}$ valuation such that $t_i^\sigma =
d_i \in {\cal D}$ and assume that ${\cal SL}$  maps $n$-ary predicates
$p$ to relations $R \subset {\cal D} \times \ldots \times {\cal D}$.
For any atomic formula $p^x(t_1,t_2,\ldots,t_n)$, and any stratified
valuation $\sigma$, where $x \in \{u,i,d\}$ and
$t_i$ are terms, the x-satisfiability relations are
defined as follows:

\begin{itemize}
\item $\sigma \uentails p^u(t_1,\ldots,t_n)$ iff 
$\langle d_1, \ldots, d_n \rangle \in R^u$
\item $\sigma \uentails p^i(t_1,\ldots,t_n)$ iff \\ \hspace*{1.4cm}
$\langle d_1, \ldots, d_n \rangle \in R^u \cup \overline{R^u} \cup R^i$
\item $\sigma \uentails p^d(t_1,\ldots,t_n)$ iff \\ \hspace*{1.4cm}
$\langle d_1, \ldots, d_n \rangle \in R^u \cup \overline{R^u}
\cup R^i \cup \overline{R^i} \cup R^d$
\item $\sigma \ientails p^u(t_1,\ldots,t_n)$ iff 
$\langle d_1, \ldots, d_n \rangle \in R^i$
\item $\sigma \ientails p^i(t_1,\ldots,t_n)$ iff 
$\langle d_1, \ldots, d_n \rangle \in R^i$
\item $\sigma \ientails p^d(t_1,\ldots,t_n)$ iff \\ \hspace*{1.4cm}
$\langle d_1, \ldots, d_n \rangle \in R^i \cup \overline{R^i}
\cup R^d$
\item $\sigma \dentails p^u(t_1,\ldots,t_n)$ iff 
$\langle d_1, \ldots, d_n \rangle \in R^d$
\item $\sigma \dentails p^i(t_1,\ldots,t_n)$ iff 
$\langle d_1, \ldots, d_n \rangle \in R^d$
\item $\sigma \dentails p^d(t_1,\ldots,t_n)$ iff 
$\langle d_1, \ldots, d_n \rangle \in R^d $
\end{itemize}
\end{definition}

Definition~\ref{xentailsfol} extends in a natural way to negated and
compound formulas. Having a satisfaction definition associated with
each level of strength provides a high degree of flexibility. The same
theory can be interpreted from a perspective that allows more
freedom (u-satisfaction), or from a perspective that is tighter
and that signals when some defeasible information has been cancelled (i-
and d-satisfaction).

Possible interpretations of a given set of utterances with respect to
a knowledge base are computed using an extension of the semantic
tableau method. This extension has been proved to be both sound and
complete~\cite{marcu-master94}. A partial ordering, $\leq$, determines
the set of {\em optimistic} interpretations for a theory.  An
interpretation $m_0$ is preferred to, or is more {\em optimistic} than, an
interpretation $m_1$ ($m_0 \leq m_1$) if it contains more information
and that information can be more easily updated in the future.  That means
that if an interpretation $m_0$ makes an utterance true by assigning
to a relation $R$ a defeasible status, while another interpretation
$m_1$ makes the same utterance true by assigning the same relation $R$
a stronger status, $m_0$ will be the preferred or {\em optimistic}
one, because it is as informative as $m_1$ and it allows more options
in the future ($R$ can be defeated).

Pragmatic inferences are triggered by utterances. To differentiate
between them and semantic inferences, we introduce a new quantifier,
$\forallu$, whose semantics is defined such that a pragmatic inference
of the form $(\forallu \vec{v})(\alpha_1(\vec{v}) \rightarrow
\alpha_2(\vec{v}))$ is instantiated only for those objects $\vec{t}$
from the universe of discourse that pertain to an utterance having the
form $\alpha_1(\vec{t})$.  Hence, only if the antecedent of a
pragmatic rule has been uttered can that rule be applied. A
meta-logical construct $uttered$ applies to the logical translation of
utterances.  This theory yields the following definition:

\begin{definition} \label{presupposition-def}
Let $\Phi$ be a theory described in terms of stratified first-order
logic that appropriately formalizes the semantics of lexical items and
the necessary conditions that trigger pragmatic inferences. The
semantics of lexical terms is formalized using the quantifier
$\forall$, while the necessary conditions that pertain to pragmatic
inferences are captured using $\forallu$.  Let $uttered(u)$ be the
logical translation of a given utterance or set of utterances. We say
that utterance $u$ \ul{\em pragmatically implicates} $p$ if and only
if $p^d$ or $p^i$ is derived using pragmatic inferences in at least
one optimistic model of the theory $\Phi \cup uttered(u)$, and if $p$
is not cancelled by any stronger information ($\neg p^u, \neg p^i, \neg
p^d$) in any optimistic model schema of the
theory. Symmetrically, one can define what a negative pragmatic
inference is.  In both cases, $\Phi \cup uttered(u)$ is u-consistent.
\end{definition}

\subsection{The algorithm}

Our algorithm, described in detail by
Marcu~\shortcite{marcu-master94}, takes as input a set of
first-order stratified formulas $\Phi$ that represents an adequate
knowledge base that expresses semantic knowledge and the necessary
conditions for triggering pragmatic inferences, and the translation of
an utterance or set of utterances $uttered(u)$. The algorithm builds
the set of all possible interpretations for a given utterance, using a
generalization of the semantic tableau technique.  The model-ordering
relation filters the {\em optimistic} interpretations. Among them, the
defeasible inferences that have been triggered on pragmatic grounds
are checked to see whether or not they are cancelled in any optimistic
interpretation.  Those that are not cancelled are labelled as
pragmatic inferences for the given utterance or set of utterances.

\section{A set of examples}

We present a set of examples that covers a representative group of
pragmatic inferences. In contrast with most other approaches, we
provide a consistent methodology for computing these inferences and
for determining whether they are cancelled or not for all possible
configurations: simple and complex utterances and sequences of
utterances.

\subsection{Simple pragmatic inferences}

\subsubsection{Lexical pragmatic inferences}

A factive such as the verb {\em regret} presupposes its complement, but
as we have seen, in positive environments, the presupposition is
stronger: it is
acceptable to defeat a presupposition triggered in a negative
environment~\sent{regret-not-cancel}, but is infelicitous to defeat
one that belongs to a positive environment~\sent{regret-yes-cancel}.
Therefore, an appropriate formalization of utterance~\sent{regret-not}
and the requisite pragmatic knowledge will be as shown in~\sent{regret-eq}.
\example{John does not regret that Mary came to the party. \name{regret-not}}
\example{
$
\left\{
\begin{array}{l}
uttered(\neg regrets^u(john, \\
\hspace*{3.0cm} come(mary,party))) \\
(\forallu x,y,z)(regrets^u(x,come(y,z)) \rightarrow \\ \hspace*{1.5cm}
come^i(y,z)) \\
(\forallu x,y,z)(\neg regrets^u(x, come(y,z)) \rightarrow  \\
\hspace*{1.5cm} come^d(y,z)) \\
\end{array}
\right.
$ \name{regret-eq}
}

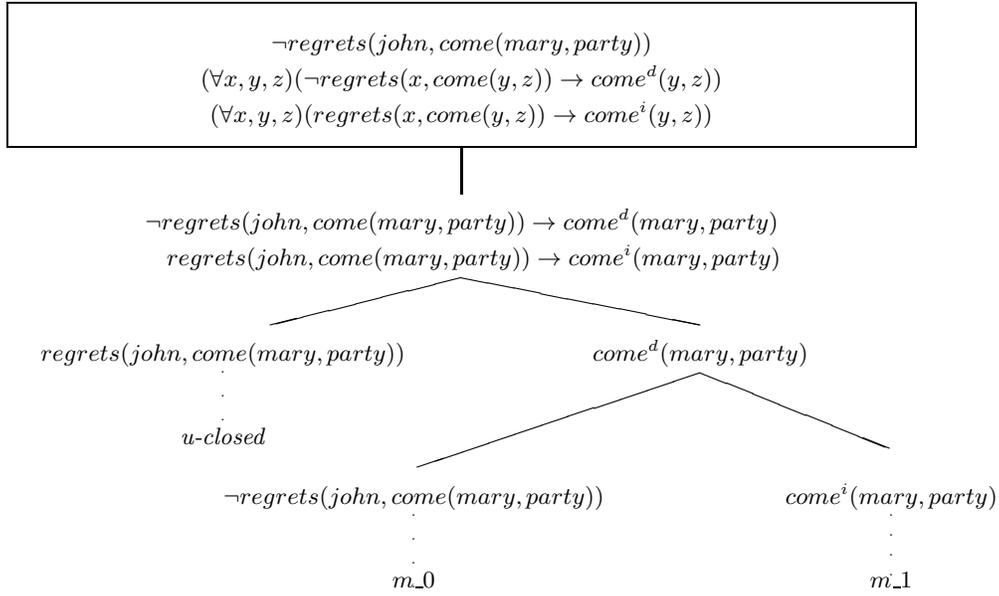
\begin{figure*}
\centering
\setlength{\unitlength}{0.012500in}%
\begingroup\makeatletter\ifx\SetFigFont\undefined
\def\x#1#2#3#4#5#6#7\relax{\def\x{#1#2#3#4#5#6}}%
\expandafter\x\fmtname xxxxxx\relax \def\y{splain}%
\ifx\x\y   
\gdef\SetFigFont#1#2#3{%
  \ifnum #1<17\tiny\else \ifnum #1<20\small\else
  \ifnum #1<24\normalsize\else \ifnum #1<29\large\else
  \ifnum #1<34\Large\else \ifnum #1<41\LARGE\else
     \huge\fi\fi\fi\fi\fi\fi
  \csname #3\endcsname}%
\else
\gdef\SetFigFont#1#2#3{\begingroup
  \count@#1\relax \ifnum 25<\count@\count@25\fi
  \def\x{\endgroup\@setsize\SetFigFont{#2pt}}%
  \expandafter\x
    \csname \romannumeral\the\count@ pt\expandafter\endcsname
    \csname @\romannumeral\the\count@ pt\endcsname
  \csname #3\endcsname}%
\fi
\fi\endgroup
\begin{picture}(465,245)(2,585)
\thinlines
\put( 30,770){\framebox(380,60){}}
\put(220,770){\line( 0,-1){ 20}}
\put(220,715){\line( 5,-1){100}}
\put(220,715){\line(-4,-1){ 80}}
\multiput(120,675)(0.00000,-10.00000){3}{\makebox(0.1111,0.7778){\SetFigFont{5}{6}{rm}.}}
\put(320,675){\line(-3,-1){118.500}}
\put(320,675){\line( 5,-2){ 79.310}}
\multiput(200,615)(0.00000,-10.00000){4}{\makebox(0.1111,0.7778){\SetFigFont{5}{6}{rm}.}}
\multiput(400,615)(0.00000,-8.33333){4}{\makebox(0.1111,0.7778){\SetFigFont{5}{6}{rm}.}}
\put(220,810){\makebox(0,0)[b]{\smash{\SetFigFont{12}{14.4}{rm}{\small $\neg
regrets(john,come(mary,party))$}}}}
\put(220,795){\makebox(0,0)[b]{\smash{\SetFigFont{12}{14.4}{rm}{\small
$(\forall x,y,z)(\neg regrets(x,come(y,z)) \rightarrow come^d(y,z))$}}}}
\put(220,780){\makebox(0,0)[b]{\smash{\SetFigFont{12}{14.4}{rm}{\small
$(\forall x,y,z)(regrets(x,come(y,z)) \rightarrow come^i(y,z))$}}}}
\put(220,735){\makebox(0,0)[b]{\smash{\SetFigFont{12}{14.4}{rm}{\small $\neg
regrets(john,come(mary,party)) \rightarrow come^d(mary,party)$}}}}
\put(225,720){\makebox(0,0)[b]{\smash{\SetFigFont{12}{14.4}{rm}{\small
$regrets(john,come(mary,party)) \rightarrow come^i(mary,party)$}}}}
\put(120,680){\makebox(0,0)[b]{\smash{\SetFigFont{12}{14.4}{rm}{\small
$regrets(john,come(mary,party))$}}}}
\put(120,645){\makebox(0,0)[b]{\smash{\SetFigFont{12}{14.4}{rm}{\small {\em
u-closed}}}}}
\put(320,680){\makebox(0,0)[b]{\smash{\SetFigFont{12}{14.4}{rm}{\small
$come^d(mary,party)$}}}}
\put(200,620){\makebox(0,0)[b]{\smash{\SetFigFont{12}{14.4}{rm}{\small $\neg
regrets(john,come(mary,party))$}}}}
\put(200,585){\makebox(0,0)[b]{\smash{\SetFigFont{12}{14.4}{rm}{\small
$m\_0$}}}}
\put(400,620){\makebox(0,0)[b]{\smash{\SetFigFont{12}{14.4}{rm}{\small
$come^i(mary,party)$}}}}
\put(400,585){\makebox(0,0)[b]{\smash{\SetFigFont{12}{14.4}{rm}{\small
$m\_1$}}}}
\end{picture}

\caption{Stratified tableau for {\em John does not regret that Mary
came to the party.}}
\label{regrets1}
\end{figure*}

\begin{figure*} 
\centering
{\small \noindent
\begin{center} \begin{tabular}{|l|l|l|l|} \hline \hline
{\em Schema \#} & {\em Indefeasible} & {\em Infelicitously}  & {\em
Felicitously} \\
 & & \hspace*{2mm} {\em defeasible} & \hspace*{2mm} {\em defeasible} \\ \hline
$m_0$ & $\neg regrets^u(john,come(mary,party)$      & &       \\
      &  &  & $come^d(mary,party)$ \\ \hline
$m_1$ & $\neg regrets^u(john,come(mary,party)$      & &       \\
      &  & $come^i(mary,party)$  & $come^d(mary,party)$ \\ \hline
\end{tabular} \end{center}
}
\caption{Model schemata for {\em John does not regret that Mary
came to the party.}}
\label{regrets1-models}
\end{figure*}

\begin{figure*} 
\centering
{\small
\begin{center} \begin{tabular}{|l|l|l|l|} \hline \hline
{\em Schema \#} & {\em Indefeasible} & {\em Infelicitously}  & {\em
Felicitously} \\
 & & \hspace*{2mm} {\em defeasible} & \hspace*{2mm} {\em defeasible}
\\ \hline
$m_0$   & $went^u(some(boys),theatre)$ & &  \\
        &  & & $\neg went^d(most(boys),theatre)$ \\
        &  & & $\neg went^d(many(boys),theatre)$ \\
        & $\neg went^u(all(boys),theatre)$  & & $\neg
went^d(all(boys),theatre)$  \\ \hline
\end{tabular} \end{center}
}
\caption{Model schema for {\em John says that some of the boys went to
the theatre.}}
\label{some-1}
\end{figure*}

\begin{figure*} 
\centering
{\small
\begin{center} \begin{tabular}{|l|l|l|l|} \hline \hline
{\em Schema \#} & {\em Indefeasible} & {\em Infelicitously}  & {\em
Felicitously} \\
 & & \hspace*{2mm} {\em defeasible} & \hspace*{2mm} {\em defeasible}
\\ \hline
$m_0$   & $went^u(some(boys),theatre)$ & &  \\
        & $went^u(most(boys),theatre)$  & & $\neg
went^d(most(boys),theatre)$ \\
        & $went^u(many(boys),theatre)$ & & $\neg
went^d(many(boys),theatre)$ \\
        & $went^u(all(boys),theatre)$  & & $\neg
went^d(all(boys),theatre)$  \\ \hline
\end{tabular} \end{center}
}
\caption{Model schema for {\em John says that some of the boys went to
the theatre.  In fact all of them went to the theatre.}}
\label{some-2}
\end{figure*}
\noindent The stratified semantic tableau that corresponds to
theory~\sent{regret-eq} is given in figure~\ref{regrets1}.
The tableau yields two model schemata (see
figure~\ref{regrets1-models}); in both of them, it is defeasibly
inferred that {\em Mary came to the party}. The model-ordering
relation $\leq$ establishes $m_0$ as the optimistic model for the
theory because it contains as much information as $m_1$ and is easier
to defeat. Model $m_0$ explains why {\em Mary came to the party} is a
presupposition for utterance~\sent{regret-not}.

\subsubsection{Scalar implicatures}

Consider utterance~\sent{scalar-ex}, and its
implicatures~\sent{scalar-ex-implic}.
\example{John says that some of the boys went to the theatre.
\name{scalar-ex}}
\example{Not \{many/most/all\} of the boys went to the theatre.
\name{scalar-ex-implic}}
An appropriate formalization is given in~\sent{scal-implic1}, where the
second formula captures the defeasible scalar
implicatures and the third formula reflects the relevant semantic
information for {\em all}.
\example{
$
\left\{
\begin{array}{l}
uttered(went(some(boys), theatre)) \\
went^u(some(boys),theatre) \rightarrow \\
\hspace*{1.3cm} (\neg went^d(many(boys),theatre) \wedge \\
\hspace*{1.3cm} \neg went^d(most(boys),theatre) \wedge \\
\hspace*{1.3cm} \neg went^d(all(boys),theatre)) \\
went^u(all(boys),theatre) \rightarrow \\
\hspace*{1.3cm} (went^u(most(boys),theatre) \wedge \\
\hspace*{1.3cm}  went^u(many(boys),theatre) \wedge \\
\hspace*{1.3cm} went^u(some(boys),theatre)) \\
\end{array}
\right.
$ \name{scal-implic1}
}
The theory provides one optimistic model schema (figure~\ref{some-1})
that reflects the expected pragmatic inferences, i.e., {\em (Not most/Not
many/Not all) of the boys went to the theatre.}

\subsubsection{Simple cancellation}

Assume now, that after a moment of thought, the same person utters:
\example{ John says that some of the boys went to the theatre. In fact
  all of them went to the theatre.}
By adding the extra utterance to the initial
theory~\sent{scal-implic1}, $uttered(went(all(boys), theatre))$, one
would obtain one optimistic model schema in which the conventional
implicatures have been cancelled (see figure~\ref{some-2}).

\subsection{Complex utterances}

The Achilles heel for most theories of presupposition  has been their
vulnerability to the projection problem.  Our solution for the
projection problem does not differ from a solution for individual
utterances. Consider the following utterances and some of their
associated presuppositions~\sent{or1-pres} (the symbol $\rhd$ precedes
an inference drawn on pragmatic grounds):
\example{Either Chris is not a bachelor or he regrets that Mary came
  to the party. \name{or1-not}}
\example{Chris is a bachelor or a spinster. \name{or1-yes}}
\example{$\rhd$ Chris is a (male) adult. \name{or1-pres}}
{\em Chris is not a bachelor} presupposes that {\em Chris is a male
adult}; {\em Chris regrets that Mary came to the party} presupposes
that {\em Mary came to the party}. There is no contradiction between
these two presuppositions, so one would expect a conversant to infer
both of them if she hears an utterance such
as~\sent{or1-not}. However, when one examines
utterance~\sent{or1-yes}, one observes immediately that there is a
contradiction between the presuppositions carried by the individual
components. Being a bachelor presupposes that {\em Chris is a male},
while being a spinster presupposes that {\em Chris is a
female}. Normally, we would expect a conversant to notice this
contradiction and to drop each of these elementary presuppositions
when she interprets \sent{or1-yes}.

We now study how stratified logic and  the model-ordering
relation capture one's intuitions.

\subsubsection{{\em Or} --- non-cancellation}

An appropriate formalization for utterance~\sent{or1-not} and the
necessary semantic and pragmatic knowledge is given in~\sent{or1-not-eq}.
\example{
$
\left\{
\begin{array}{l}
uttered(\neg bachelor(Chris) \vee \\
\hspace*{1.4cm} regret(Chris, come(Mary,party))) \\
(\neg bachelor^u(Chris) \vee \\
\hspace*{2mm} regret^u(Chris, come(Mary,party))) \rightarrow  \\
\hspace*{5mm} \neg  (\neg bachelor^d(Chris) \wedge \\
\hspace*{5mm} regret^d(Chris,come(Mary,party))) \\
\neg male(Mary) \\
(\forall x)(bachelor^u(x) \rightarrow \\
\hspace*{5mm} male^u(x) \wedge adult^u(x) \wedge
\neg married^u(x)) \\
(\forallu x) (\neg bachelor^u(x) \rightarrow married^i(x)) \\
(\forallu x) (\neg bachelor^u(x) \rightarrow adult^d(x)) \\
(\forallu x) (\neg bachelor^u(x) \rightarrow male^d(x)) \\
(\forallu x,y,z)(\neg regret^u(x, come(y,z)) \rightarrow \\
\hspace*{1.5cm}  come^d(y,z)) \\
(\forallu x,y,z)(regret^u(x, come(y,z)) \rightarrow \\
\hspace*{1.5cm} come^i(y,z)) \\
\end{array}
\right.
$ \name{or1-not-eq}
}
Besides the translation of the utterance, the initial theory contains
a formalization of the defeasible implicature that natural disjunction
is used as an exclusive {\em or}, the knowledge that {\em Mary} is not
a name for males, the lexical semantics for the word
{\em bachelor}, and the lexical pragmatics for {\em bachelor} and {\em
regret}.  The stratified semantic tableau generates 12 model
schemata. Only four of them are kept as optimistic models for the
utterance.  The models yield {\em Mary came to the party; Chris is a
male;} and {\em Chris is an adult} as pragmatic inferences of
utterance~\sent{or1-not}.

\subsubsection{{\em Or} --- cancellation}

Consider now utterance~\sent{or1-yes}.  The stratified semantic
tableau that corresponds to its logical theory yields 16 models, but
only {\em Chris is an adult} satisfies
definition~\ref{presupposition-def} and is projected as presupposition
for the utterance.

\subsection{Pragmatic inferences in sequences of utterances}
\label{section-seq}

We have already mentioned that speech repairs constitute a good
benchmark for studying the generation and cancellation of pragmatic
inferences along sequences of utterances~\cite{mcroy93}.  Suppose, for
example, that Jane has two friends --- John Smith and John Pevler ---
and that her roommate Mary has met only John Smith, a married fellow.
Assume now that Jane has a conversation with Mary in which Jane
mentions only the name John because she is not aware that Mary does
not know about the other John, who is a five-year-old boy. In this
context, it is natural for Mary to become confused and to come to wrong
conclusions. For example, Mary may reply that {\em John is not a
  bachelor}. Although this is true for both Johns, it is more
appropriate for the married fellow than for the five-year-old boy.
Mary knows that John Smith is a married male, so the utterance makes
sense for her. At this point Jane realizes that Mary misunderstands
her: all the time Jane was talking about John Pevler, the
five-year-old boy. The utterances in~\sent{utt1} constitute a possible
answer that Jane may give to Mary in order to clarify the problem.
\example{ a. No, John is not a bachelor. \\
b. I regret that you have misunderstood me. \\
c. He is only five years old. \name{utt1}}
The first utterance in the sequence presupposes~\sent{utt1-pres}.
\example{ $\rhd$ John is a male adult.
\name{utt1-pres}}
Utterance~\sent{utt1}b warns Mary that is very likely she misunderstood a
previous utterance~\sent{utt2-pres}. The warning is conveyed by
implicature.
\example{ $\rhd$ The hearer misunderstood the speaker. \name{utt2-pres}}
At this point, the hearer, Mary, starts to
believe that one of her previous utterances has been elaborated on a
false assumption, but she does not know which one. The third
utterance~\sent{utt1}c comes to clarify the issue. It explicitly
expresses that John is not an adult. Therefore, it
cancels the early presupposition~\sent{utt1-pres}:
\example{ $\rhd \! \! \! \!/$ John is an adult.
\name{utt3-pres}}
Note that there is a gap of one statement between the generation and
the cancellation of this presupposition. The behavior described is
mirrored both by our theory and our program.

\subsection{Conversational implicatures in indirect replies}

The same methodology can be applied to modeling conversational
implicatures in indirect replies \cite{green92}. Green's algorithm
makes use of discourse expectations, discourse plans, and discourse
relations. The following dialog is considered~\cite[p.~68]{green92}:
\example{Q: \hspace*{3mm} Did you go shopping? \\
         A: a. My car's not running. \\
            \hspace*{4mm} b. The timing belt broke. \\
            \hspace*{4mm} c. (So) I had to take the bus. \name{dialog-1}}

\noindent Answer~\sent{dialog-1} conveys a ``yes'', but a reply  consisting
only of~\sent{dialog-1}a would implicate a ``no''. As Green notices,
in previous models of implicatures~\cite{gazdar79,hirschberg85},
processing~\sent{dialog-1}a will block the implicature generated
by~\sent{dialog-1}c. Green solves the problem by extending the
boundaries of the analysis to discourse units. Our approach does not
exhibit these constraints. As in the previous example, the one dealing
with a sequence of utterances, we obtain a different interpretation
after each step. When the question is asked, there is no
conversational implicature. Answer~\sent{dialog-1}a makes the
necessary conditions for implicating ``no'' true, and the implication
is computed. Answer~\sent{dialog-1}b reinforces a previous condition.
Answer~\sent{dialog-1}c makes the preconditions for implicating a
``no'' false, and the preconditions for implicating a ``yes'' true.
Therefore, the implicature at the end of the dialogue is that the
conversant who answered went shopping.

\section{Conclusions}

Unlike most research in pragmatics that focuses on certain types of
presuppositions or implicatures, we provide a global framework in
which one can express all these types of pragmatic inferences. Each
pragmatic inference is associated with a set of necessary conditions
that may trigger that inference.  When such a set of conditions is
met, that inference is drawn, but it is assigned a defeasible status.
An extended definition of satisfaction and a notion of ``optimism''
with respect to different interpretations yield the preferred
interpretations for an utterance or sequences of utterances. These
interpretations contain the pragmatic inferences that have not been
cancelled by context or conversant's knowledge, plans, or intentions.
The formalism yields an algorithm that has been implemented in Common
Lisp with Screamer. This algorithm computes uniformly pragmatic
inferences that are associated with simple and complex utterances and
sequences of utterances, and allows cancellations of pragmatic
inferences to occur at any time in the discourse.

\vspace{6mm}

\noindent{\large\bf Acknowledgements}

\vspace{3mm}

\noindent This research was supported in part by a grant from the Natural
Sciences and Engineering Research Council of Canada.

\pagebreak


\begin{thebibliography}{}

\bibitem[\protect\citename{Frege}1892]{frege92}
G.~Frege.
\newblock 1892.
\newblock {\"{U}ber} sinn und bedeutung.
\newblock {\em Zeitschrift {f\"{u}r} Philos. und Philos. Kritik}, 100:373--394.
\newblock reprinted as: On Sense and Nominatum, In Feigl H. and Sellars W.,
  editors, {\em Readings in Philosophical Analysis}, pages 85--102,
  Appleton-Century-Croft, New York, 1947.

\bibitem[\protect\citename{Gazdar}1979]{gazdar79}
G.J.M. Gazdar.
\newblock 1979.
\newblock {\em Pragmatics: Implicature, Presupposition, and Logical Form}.
\newblock Academic Press.

\bibitem[\protect\citename{Green and Carberry}1994]{green94}
N.~Green and S.~Carberry.
\newblock 1994.
\newblock A hybrid reasoning model for indirect answers.
\newblock In {\em Proceedings 32nd Annual Meeting of the Association for
  Computational Linguistics}, pages 58--65.

\bibitem[\protect\citename{Green}1990]{green90}
N.~Green.
\newblock 1990.
\newblock Normal state implicature.
\newblock In {\em Proceedings 28th Annual Meeting of the Association for
  Computational Linguistics}, pages 89--96.

\bibitem[\protect\citename{Green}1992]{green92}
N.~Green.
\newblock 1992.
\newblock Conversational implicatures in indirect replies.
\newblock In {\em Proceedings 30th Annual Meeting of the Association for
  Computational Linguistics}, pages 64--71.

\bibitem[\protect\citename{Hirschberg}1985]{hirschberg85}
J.B. Hirschberg.
\newblock 1985.
\newblock A theory of scalar implicature.
\newblock Technical Report MS-CIS-85-56, Department of Computer and Information
  Science, University of Pennsylvania.
\newblock Also published by Garland Publishing Inc., 1991.

\bibitem[\protect\citename{Hirst \bgroup et al.\egroup }1994]{hirst94}
G.~Hirst, S.~McRoy, P.~Heeman, P.~Edmonds, and D.~Horton.
\newblock 1994.
\newblock Repairing conversational misunderstandings and non-understandings.
\newblock {\em Speech Communication}, 15:213--229.

\bibitem[\protect\citename{Hirst}1991]{hirst91}
G.~Hirst.
\newblock 1991.
\newblock Existence assumptions in knowledge representation.
\newblock {\em Artificial Intelligence}, 49:199--242.

\bibitem[\protect\citename{Karttunen and Peters}1979]{karttunen79}
L.~Karttunen and S.~Peters.
\newblock 1979.
\newblock Conventional implicature.
\newblock In Oh~C.K. and Dinneen D.A, editors, {\em Syntax and Semantics,
  Presupposition}, volume~11, pages 1--56. Academic Press.

\bibitem[\protect\citename{Karttunen}1974]{karttunen74}
L.~Karttunen.
\newblock 1974.
\newblock Presupposition and linguistic context.
\newblock {\em Theoretical Linguistics}, 1:3--44.

\bibitem[\protect\citename{Kay}1992]{kay92}
P.~Kay.
\newblock 1992.
\newblock The inheritance of presuppositions.
\newblock {\em Linguistics \& Philosophy}, 15:333--379.

\bibitem[\protect\citename{Marcu}1994]{marcu-master94}
D.~Marcu.
\newblock 1994.
\newblock A formalism and an algorithm for computing pragmatic inferences and
  detecting infelicities.
\newblock Master's thesis, Dept. of Computer Science, University of Toronto,
  September.
\newblock Also published as Technical Report CSRI-309, Computer Systems
  Research Institute, University of Toronto.

\bibitem[\protect\citename{Marcu and Hirst}1994]{marcu-tilburg94}
D.~Marcu and G.~Hirst.
\newblock 1994.
\newblock An implemented formalism for computing linguistic presuppositions and
  existential commitments.
\newblock In H.~Bunt, R.~Muskens, and G.~Rentier, editors, {\em International
  Workshop on Computational Semantics}, pages 141--150, December.

\bibitem[\protect\citename{McRoy and Hirst}1993]{mcroy93}
S.~McRoy and G.~Hirst.
\newblock 1993.
\newblock Abductive explanation of dialogue misunderstandings.
\newblock In {\em Proceedings, 6th Conference of the European Chapter of the
  Association for Computational Linguistics}, pages 277--286, April.

\bibitem[\protect\citename{Mercer}1987]{mercerphd}
R.E. Mercer.
\newblock 1987.
\newblock {\em A Default Logic Approach to the Derivation of Natural Language
  Presuppositions}.
\newblock {Ph.D.} thesis, Department of Computer Science, University of British
  Columbia.

\bibitem[\protect\citename{Quine}1949]{quine49}
W.V.O. Quine.
\newblock 1949.
\newblock Designation and existence.
\newblock In Feigl H. and Sellars W., editors, {\em Readings in Philosophical
  Analysis}, pages 44--51. Appleton-Century-Croft, New York.

\bibitem[\protect\citename{Reiter}1980]{reiter80}
R.~Reiter.
\newblock 1980.
\newblock A logic for default reasoning.
\newblock {\em Artificial Intelligence}, 13:81--132.

\bibitem[\protect\citename{Russell}1905]{russell05}
B.~Russell.
\newblock 1905.
\newblock On denoting.
\newblock {\em Mind n.s.}, 14:479--493.
\newblock reprinted in: Feigl H. and Sellars W. editors, {\em Readings in
  Philosophical Analysis}, pages 103--115. Appleton-Century-Croft, New York,
  1949.

\bibitem[\protect\citename{Sandt}1992]{sandt92}
{R.A. van der} Sandt.
\newblock 1992.
\newblock Presupposition projection as anaphora resolution.
\newblock {\em Journal of Semantics}, 9:333--377.

\bibitem[\protect\citename{Siskind and McAllester}1993]{siskind93a}
J.M. Siskind and D.A. McAllester.
\newblock 1993.
\newblock Screamer: A portable efficient implementation of nondeterministic
  Common Lisp.
\newblock Technical Report IRCS-93-03, University of Pennsylvania, Institute
  for Research in Cognitive Science, July 1.

\bibitem[\protect\citename{Weischedel}1979]{weischedel79}
R.M. Weischedel.
\newblock 1979.
\newblock A new semantic computation while parsing: Presupposition and
  entailment.
\newblock In Oh~C.K. and Dinneen D.A, editors, {\em Syntax and Semantics,
  Presupposition}, volume~11, pages 155--182. Academic Press.

\bibitem[\protect\citename{Zeevat}1992]{zeevat92}
H.~Zeevat.
\newblock 1992.
\newblock Presupposition and accommodation in update semantics.
\newblock {\em Journal of Semantics}, 9:379--412.

\end{thebibliography}

\end{document}